\documentclass[sigconf]{acmart}

\usepackage{subfig}
\usepackage{amsfonts,amssymb} 
\usepackage{amsmath}
\usepackage{amsthm}
\usepackage{bm}
\usepackage{multirow}
\usepackage{makecell}
\usepackage{arydshln}
\usepackage{booktabs}
\usepackage{multicol}
\usepackage{color}
\usepackage{caption}
\definecolor{darkgreen}{RGB}{10,105,6}
\definecolor{navy}{RGB}{176,7,14}
\definecolor{dblue}{RGB}{30,14,255}

\AtBeginDocument{%
  }

\setcopyright{acmcopyright}
\copyrightyear{2023}
\acmYear{2023}
\acmDOI{XXXXXXX.XXXXXXX}

\acmConference[MM '23]{Proceedings of the 31th ACM International Conference on Multimedia}{October 28--Novmber 03,
  2023}{Ottawa, Canada}
\acmBooktitle{Proceedings of the 31th ACM International Conference on Multimedia (MM '23), October 28--Novmber 03, 2023, Ottawa, Canada}
\acmPrice{15.00}
\acmISBN{978-1-4503-XXXX-X/18/06}

\renewcommand\footnotetextcopyrightpermission[1]{}

\title{Cross-Architecture Distillation for Face Recognition}

\author{
    Weisong Zhao \textsuperscript{\rm1,3},
    Xiangyu Zhu \textsuperscript{\rm2,4\authornotemark[1]},
    Zhixiang He \textsuperscript{\rm5},
    Xiao-Yu Zhang \textsuperscript{\rm1,3},
    Zhen Lei \textsuperscript{\rm2,4,6} 
}

\affiliation{
    \textsuperscript{\rm 1}Institute of Information Engineering, Chinese Academy of Sciences, Beijing, China\\
\textsuperscript{\rm 2} CBSR\&NLPR, Institute of Automation, Chinese Academy of Sciences, Beijing, China\\
\textsuperscript{\rm 3} School of Cyber Security, University of Chinese Academy of Sciences, Beijing, China\\
\textsuperscript{\rm 4} School of Artificial Intelligence, University of Chinese Academy of Sciences \city{Beijing} \country{China}\\
\textsuperscript{\rm 5} China Telecom Corporation Ltd. Data \& AI Technology Company\\
\textsuperscript{\rm 6} Centre for Artificial Intelligence and Robotics, Hong Kong Institute of Science \& Innovation,\\
Chinese Academy of Sciences, Hong Kong, China\\
   \{zhaoweisong, zhangxiaoyu\}@iie.ac.cn, hezx3@chinatelecom.cn, \{xiangyu.zhu, zlei\}@nlpr.ia.ac.cn
}

\renewcommand{\shortauthors}{Zhao et al.}
\begin{abstract}
Transformers have emerged as the superior choice for face recognition tasks, but their insufficient platform acceleration hinders their application on mobile devices. In contrast, Convolutional Neural Networks (CNNs) capitalize on hardware-compatible acceleration libraries. Consequently, it has become indispensable to preserve the distillation efficacy when transferring knowledge from a Transformer-based teacher model to a CNN-based student model, known as Cross-Architecture Knowledge Distillation (CAKD). Despite its potential, the deployment of CAKD in face recognition encounters two challenges: 1) the teacher and student share disparate spatial information for each pixel, obstructing the alignment of feature space, and 2) the teacher network is not trained in the role of a teacher, lacking proficiency in handling distillation-specific knowledge. To surmount these two constraints, 1) we first introduce a Unified Receptive Fields Mapping module (URFM) that maps pixel features of the teacher and student into local features with unified receptive fields, thereby synchronizing the pixel-wise spatial information of teacher and student. Subsequently, 2) we develop an Adaptable Prompting Teacher network (APT) that integrates prompts into the teacher, enabling it to manage distillation-specific knowledge while preserving the model's discriminative capacity. Extensive experiments on popular face benchmarks and two large-scale verification sets demonstrate the superiority of our method.
\end{abstract}



\received{20 February 2007}
\received[revised]{12 March 2009}
\received[accepted]{5 June 2009}

\begin{document}
\maketitle
\section{Introduction}
Face recognition has attained tremendous success in various application areas \cite{liziqing, leizhen, spl}. However, compact yet discriminative face recognition models are highly desirable due to the proliferation of identification systems on mobile and peripheral devices \cite{ekd}. Despite the variant proposals of enhanced neural network designs, there remains an immense performance disparity between these compressed networks and the heavy networks with millions of parameters. A natural option is to optimize neural network architectures for mobile devices, e.g., MobileFaceNet \cite{mobilefacenet}, and MobileNetV3 \cite{mobilenetv3}. However, discriminative networks always benefit from a large modeling capacity, which is time- and labor-intensive. Knowledge distillation (KD) refers to the vanilla method for enhancing the performance of light models \cite{kd, fitnet}. A typical scenario involves distilling either the intermediate features or subsequent logits from a strong teacher neural network to a compact student network, aiming at substantially improving the performance of the student model. Nevertheless, existing KD techniques primarily focus on homologous-architecture distillation, i.e., CNN to CNN.
\begin{figure}[t]
\centering
\subfloat[\centering{Performance variation on LFW, CFP-FP, CPLFW, AgeDB and CALFW}]{
\centering
\label{fig1:a}
\includegraphics[width=0.49\linewidth]{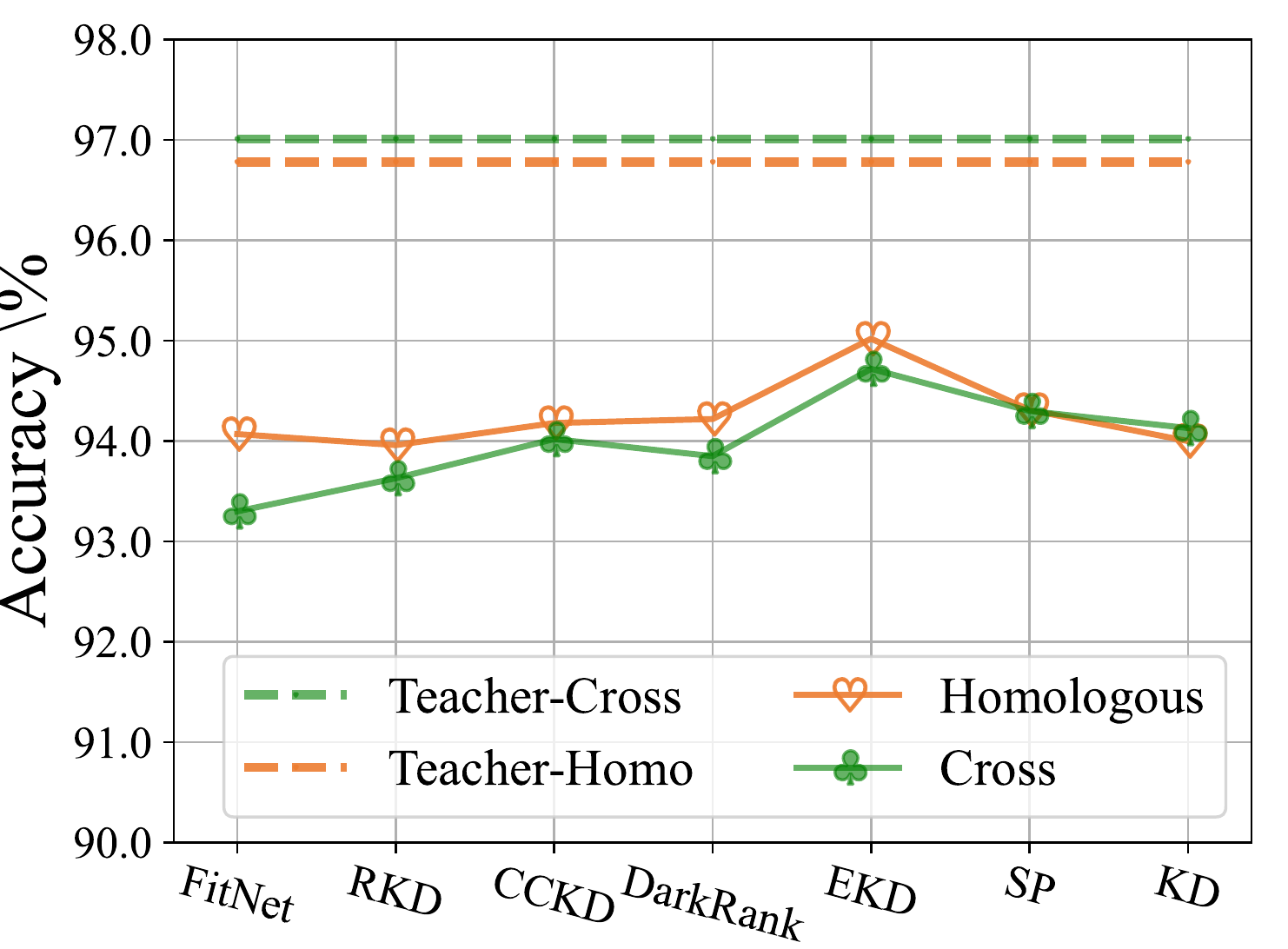}
}
\centering
\subfloat[\centering{Performance variation on IJB-B and IJB-C datasets}]{
\centering
\label{fig1:b}
\includegraphics[width=0.49\linewidth]{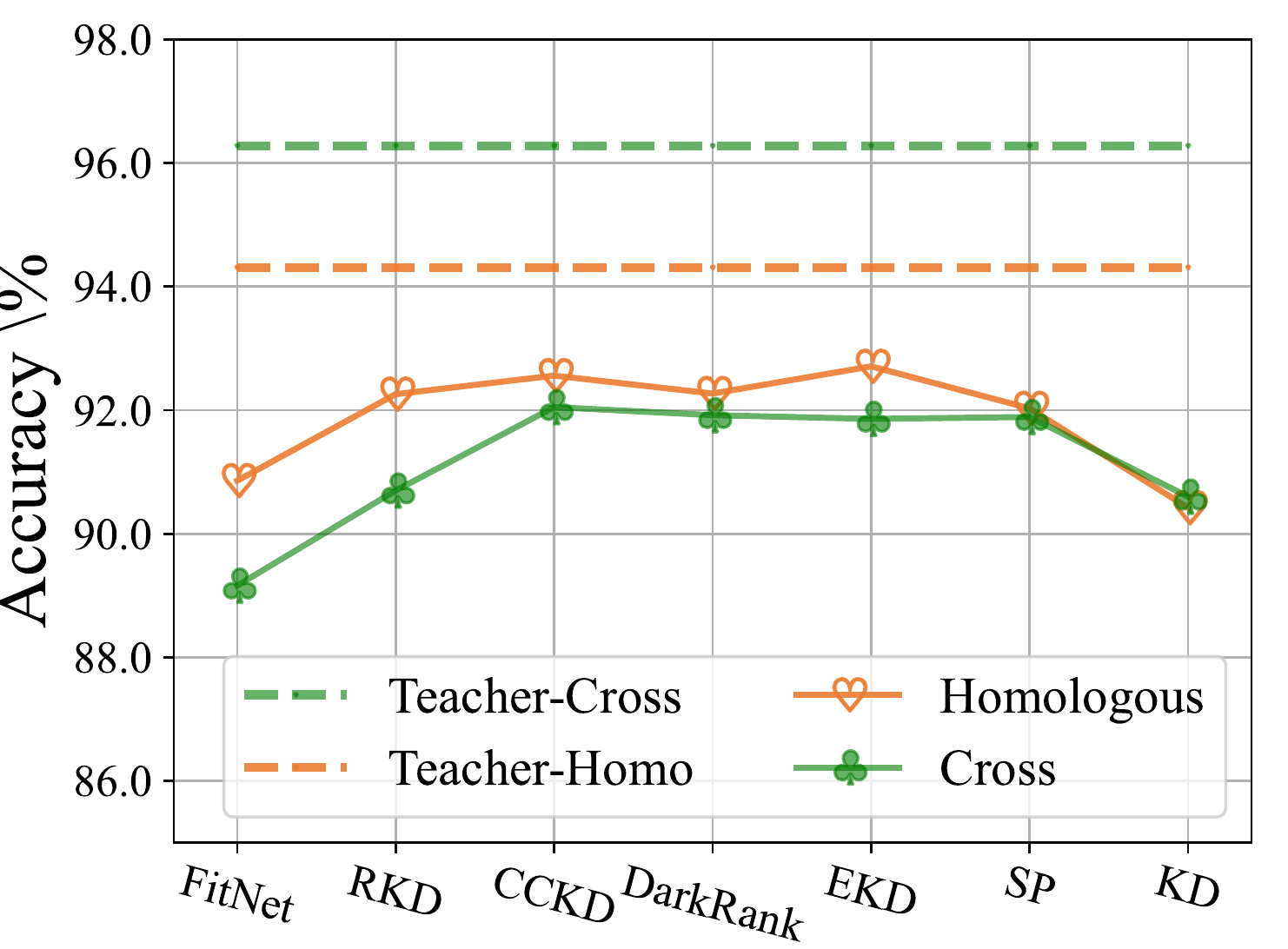}
}
\caption{Existing KD methods suffer from performance degradation in cross-architecture distillation compared to the homologous-architecture distillation. With the student network identified as MobileFacenet \cite{mobilefacenet}, we adopt IResNet-50 \cite{arcface} as the teacher for homologous-architecture distillation and Swin-S as the teacher for cross-architecture distillation. Then, we evaluate the performance variation of students with different KD methods \cite{fitnet, kd, sp, darkrank, cckd, rkd, ekd} in both scenarios by: (a) the average accuracy on the five popular face benchmarks \cite{lfw, cfp-fp, cplfw, agedb, calfw}, and (b) the average TPR@FAR=1e-4 on IJB-B \cite{ijb-b} and IJB-C \cite{ijb-c}. Practical application requires a solution to transfer knowledge from Transformer to CNN, which serves as the primary focus of this study.}
\vspace{-0.3cm}
\label{fig1}
\end{figure}

Recently, Transformers have demonstrated exceptional capabilities in various vision tasks \cite{vit, swin, detect}. Nonetheless, their high computational requirements and insufficient support for platform acceleration have hindered their deployment on mobile devices.  On the other hand, CNNs have undergone significant development in recent years, with hardware-friendly acceleration libraries such as CUDA \cite{cuda} and TensorRT \cite{tensorrt} rendering them suitable for both servers and mobile devices. Consequently, considering the exceptional modelling capacity of Transformers and the compatibility of CNNs, it has become a prevalent practice to employ a Transformer as a teacher network and maintain CNN as a student network for KD. However, current KD methods concentrate on homologous-architecture distillation and overlook the architectural gap between teacher and student networks, leading to inferior performance of Transformer to CNN compared with that of CNN to CNN.
Therefore, we probe the implication of the architecture gap on knowledge distillation in face recognition. Specifically, we first trained an IResNet-50 \cite{arcface} on MS1M-V2 \cite{arcface} as a teacher network under the homologous-architecture scenario, followed by training a Swin-S \cite{swin} as a teacher network under the cross-architecture scenario. It is worth noting that Swin-S has a slight performance improvement over the IResNet-50. For the student network, we choose MobileFaceNet \cite{mobilefacenet} as the backbone. We reproduce the canonical knowledge distillation methods \cite{fitnet, kd, sp, darkrank, cckd, rkd, ekd} in face recognition in the homologous- and cross-architecture settings, respectively. We calculate the performance variation of different methods from homologous- to cross-architecture scenarios, as shown in Fig. \ref{fig1}. Most methods suffer from performance degradation in cross-architecture scenarios. However, we believe Cross-architecture knowledge distillation is still effective in face recognition due to the highly organized face structure.
\begin{figure}[t]
\centering 
\vspace{0.1cm}
\includegraphics[width=\columnwidth]{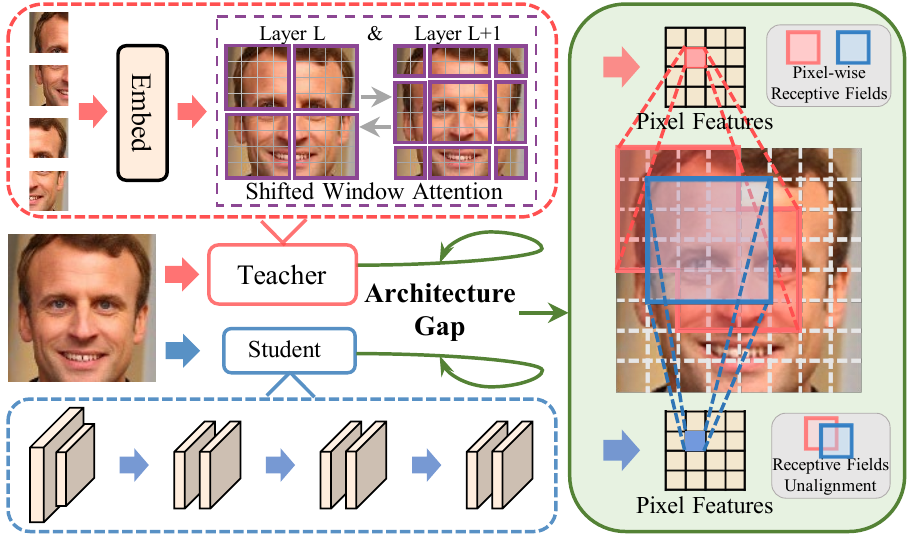}
\caption{Illustration of the receptive fields of a pixel feature for teacher (Swin) and student (CNN). There exists a theoretical receptive field unalignment between teacher and student due to the architectural difference.}

\label{fig2}
\end{figure}

In this paper, we find that the deployment of cross-architecture knowledge distillation in face recognition encounters two major challenges. First, as illustrated in Fig. \ref{fig2}, there is a significant architecture gap between teacher and student networks in terms of pixel-wise receptive fields, i.e., the teacher network adopts shifted window attention \cite{swin} and the student utilizes conventional convolution operations. To demonstrate this, We visualize ERF \cite{erf} of pixel-wise receptive fields for the teacher and student. As illustrated in Sec \ref{section:numofprompt}, the teacher and student share disparate pixel-wise spatial information. Second, the teacher network is not trained in the role of a teacher, lacking awareness of managing distillation-specific knowledge. The challenge lies in developing an auxiliary module that enables the teacher to manage distillation-specific knowledge while preserving its discriminative capacity.

To address the aforementioned challenges, we first introduce a Unified Receptive Fields Mapping module (URFM) designed to map pixel features of the teacher and student models into local features with congruent receptive fields. To achieve this, we utilize learnable local centers as the query embedding, supplemented with facial positional encoding to synchronize the receptive fields of the pixel features in both teacher and student. Additionally, recent research explores the feasibility of prompts in visual recognition and continual learning \cite{vpt, continual1, continual2}. In this paper, we investigate the applicability of prompts in KD, allowing the teacher to optimize during distillation. Specifically, we develop an Adaptable Prompting Teacher network (APT) that integrates prompts into the teacher, enabling it to manage distillation-specific knowledge.

In summary, the contributions of this paper include:

\begin{itemize}
\item We propose a novel module called Unified Receptive Fields Mapping (URFM) that maps pixel-wise features to local features with unified receptive fields. In URFM module, we exploit learnable local centers as the query embedding on which we supplement a facial positional encoding with the facial key points to synchronize the pixel-wise receptive fields of teacher and student networks.
\item We introduce Adaptable Prompting Teacher network (APT) that supplements learnable prompts in the teacher, enalbing it to manage distillation-specific knowledge while preserving model's discriminative capacity. We further propose to adapt the model's adaptable capacity by altering the number of prompts. To the best of our knowledge, we are the first to explore the feasibility of prompts in KD.
\item The extensive experiments on popular face recognition benchmarks demonstrate the superiority of the proposed method over the state-of-the-art methods.
\end{itemize}

\section{Related Work}
\subsection{Face Recognition}
Face recognition (FR) is a demanding computer vision task that seeks to identify or authenticate a person's identity based on their facial features. A crucial component of face recognition systems is the loss function, responsible for measuring the similarity or dissimilarity between face embeddings. Two primary loss functions are employed in face recognition: verification loss and softmax-based loss. The former optimizes pairwise Euclidean distance in feature space using contrastive loss \cite{contrastive1,contrastive2} or differentiates positive pairs from negative pairs by applying a distance margin through triplet loss \cite{facenet,tripletnet}. The latter is extensively adopted by state-of-the-art deep face recognition methods.
Softmax loss functions combined with heavy neural networks are demonstrated to obtain satisfactory performance \cite{arcface}. Various methods have been proposed to learn features with angular discrimination. SphereFace \cite{sphereface} introduces the angular SoftMax function (i.e., A-SoftMax), adding discriminative constraints on a hypersphere manifold. CosFace \cite{cosface} further suggests a large margin cosine loss to enhance the decision margin in the angular space. ArcFace loss is designed to achieve highly discriminative features for FR by incorporating angular margin loss \cite{arcface}. CurricularFace \cite{curricularface} integrates the concept of curriculum learning into the loss function. MagFace \cite{magface} explores applying different margins based on recognizability, which incorporates substantial angular margins for elevated-norm features that exhibit a heightened level of discernibility. These loss functions differ in their approaches to optimizing intra-class compactness and inter-class separability of face embeddings. However, most of these loss functions rely on large-scale training data and high-capacity models, constraining their applicability on mobile devices.

\subsection{Knowledge Distillation}
Knowledge distillation was first proposed by \citet{kd}, who suggested transferring the softened logits (before the softmax layer) from the teacher to the student by minimizing the Kullback-Leibler divergence. A temperature factor is used to smooth the logits. In pursuit of richer representations, \citet{fitnet} proposed transferring intermediate layer features between the teacher and student networks. Subsequently, \citet{at} devised several statistical methods to emphasize the dominant areas of the feature map and disregard low-response areas as noise. \citet{crosslayer} introduced semantic calibration, enabling the student to learn from the most semantically related teacher layer. In \cite{featurematch}, feature similarities between the teacher and student networks are computed and utilized as weights to balance feature matching. However, these approaches overlook the problem of semantic mismatch, where pixels in the teacher feature map often contain more semantic information than those in the student map at the same spatial location. We note that some works \cite{interchannel, pkt, rkd, cckd, ekd} relax the spatial constraint during feature distillation. Typically, they define a relational graph or similarity matrix in the teacher network's feature space and transfer it to the student network. For instance, \citet{sp} calculates a similarity matrix, with each entry encoding the similarity between two instances. \citet{interchannel} measure the correlation between channels using inner products. These methods reduce and compress entire features to specific properties, thereby eliminating spatial information. However, existing methods predominantly focus on homologous-architecture KD, limiting their applicability in cross-architecture KD.

\subsection{Cross-architecture Knowledge Distillation}
Transformers have been applied to various computer vision tasks, such as image classification  \cite{vit}, object detection \cite{detect}, semantic segmentation \cite{seg}, face recognition \cite{face} and video understanding \cite{video}. They have shown competitive or superior performance compared to convolutional neural networks (CNNs) on many benchmarks and datasets \cite{survey}. However, Transformers are computationally expensive and hard to accelerate on different platforms, especially on mobile devices. On the other hand, CNNs have been well developed in recent years, with libraries like CUDA \cite{cuda} and TensorRT \cite{tensorrt} that make them compatible with both servers and edge devices. Therefore, a common practice is to utilize Transformer as a teacher network and CNN as a student network for KD, which can improve the student’s performance. Many existing KD methods cannot work with Transformers due to the architecture gap between Transformer and CNN \cite{accv}. Some works have studied how to distill knowledge between Transformers. For example, DeiT \cite{deit} supplements a distillation token to assist the student Transformer to learn from the teacher and the ground truth (GT). MINILM [21] focuses on distilling the self-attention information in Transformer. IR [22] transfers the internal representations (e.g., self-attention map) from the teacher to the student. However, most of these methods require similar or identical architectures for both teacher and student. To solve this problem, \citet{accv} proposes to align the attention space and feature space of teacher and student networks, assuming that they share identical spatial information for each pixel. However, we argue that this assumption does not hold. As illustrated in Fig. \ref{fig2}, there is a significant architecture gap between teacher and student networks in terms of pixel-wise receptive fields: the teacher network adopts shifted window attention \cite{swin}, while the student utilizes conventional convolution operations. In accordance with \cite{accv}, we mitigate the architecture gap by aligning attention space and feature space. Having an edge on it, we synchronize the receptive fields of pixel-wise features of student and teacher networks, which further alleviates the architecture gap.

\subsection{Prompting in Vision}
Prompting is a technique that utilizes language instruction at the beginning of the input text to assist a pre-trained language model in pre-understanding the task \cite{prompt}. GPT-3 \cite{gpt3} strongly generalizes downstream transfer learning tasks with manually chosen prompts, even in few-shot or zero-shot settings. Recent works propose to optimize the prompts as task-specific continuous vectors via gradients during fine-tuning, which is called Prompt Tuning \cite{prompttuning1, prompttuning2, prompttuning3}. It performs comparable to full fine-tuning but with hundreds of times less parameter storage. \citet{vpt} explore the generality and feasibility of visual prompting across multiple domains. \citet{continual1} probe the viability of prompts in continual learning. In this paper, we explore prompts' applicability in KD. The objective is to optimize prompts to instruct the teacher to manage distillation-specific knowledge while maintaining model adaptable capacity.

\section{Method}
In this section, we first provide an overview of the proposed method, followed by a brief introduction to the general formulation of the Adaptable Prompting Teacher network (APT). Next, we detail the design of the proposed Unified Receptive Fields Mapping module (URFM). Lastly, we introduce the implementation of Facial Positional Encoding (FPE), including two candidate metric schemes, i.e., Saliency Distance (SD) and Relative Distance (RD).
\begin{figure*}[htb]
\centering
\includegraphics[width=\linewidth]{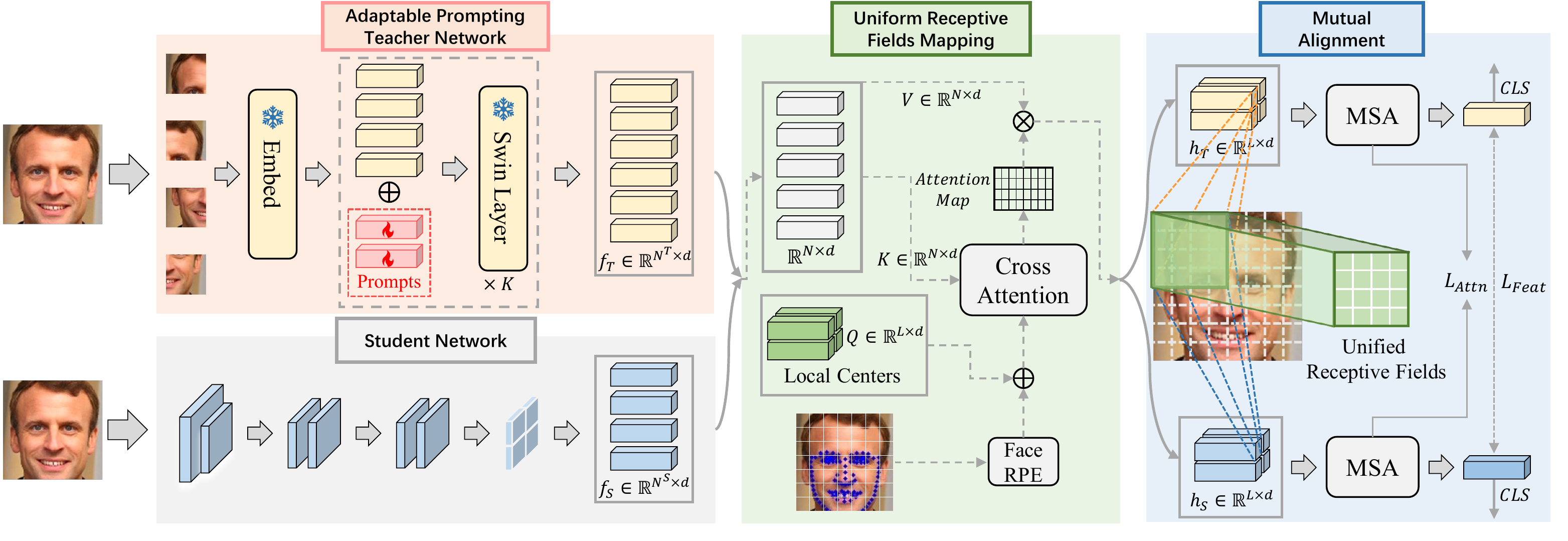}
\vspace{-0.5cm}
\caption{An overview of the proposed method encompassing Adaptable Prompting Teacher Network (APT), Unified Receptive Fields Mapping module (URFM), as well as the reciprocal alignment in feature space and attention space. For the teacher network, a facial image is initially divided into $m$ patches encoded via a linear projection. The patch embeddings are then fed to APT to produce pixel features $\bm{f}^T$ which are subsequently mapped through URFM to obtain the local features $\bm{h}^T$ with unified receptive fields. Likewise, the ultimate pixel features $\bm{f}^S$ of the student are produced after encoding of CNN layers, followed by URFM to get local features $\bm{h}^S$. We eventually execute a reciprocal alignment in the feature space and attention space.}
\label{fig3}
\vspace{-0.1cm}
\end{figure*}
\subsection{Framework}
We present the overall framework of our method in Fig. \ref{fig3}. The upper pink network and the lower gray network are the teacher and student networks, respectively. Suppose the teacher and student are Swin Transformer and convolutional neural networks, denoted by $T$ and $S$. For the teacher network (Transformer), an input facial image is resized and divided into $m$ patches $\mathbf{x}=\{x_j\in \mathbb{R}^{h\times w\times 3}\, |\,1\le j\le m\}$, which are then fed into Adaptable Prompting Teacher to get $N^T$ $d$-dimension pixel features (image tokens). $h$ and $w$ indicate the height and width of the patches. Then the pixel features $\bm{f}^T\in\mathbb{R}^{N^T\times d}$ are input into the URFM module to get local features $\bm{h}^T\in\mathbb{R}^{L\times d}$ with unified pixel receptive fields. For the student (CNN), the pixel features $\bm{f}^S\in\mathbb{R}^{N^S\times d}$ are produced after the encoding of several CNN layers, followed by URFM to get local features $\bm{h}^S  \in \mathbb{R}^{L\times d}$. $L$ and $d$ denote the number of local centers in URFM and feature dimension, respectively. Note that the values of $N^T$ and $N^S$ are commonly different, due to the distinct architecture of the teacher and student networks as well as the supplemented prompts. Finally, we conduct a mutual alignment on the attention space and feature space between teacher and student networks. The adopted classification loss is ArcFace loss \cite{arcface} and the alignment losses on attention space and feature space are formulated as follows:
\begin{equation}
\begin{aligned}
\mathcal{L}_{Attn} &= \mbox{MSE}(Attn_T, Attn_S) ,\\
\mathcal{L}_{Feat} = &\mbox{MSE}(\mbox{MSA}(\bm{h}^T)-\mbox{MSA}(\bm{h}^S)).
\end{aligned}
\end{equation}
$Attn_T$ and $Attn_S$ represent the final attention maps of the teacher and student, respectively. $\mbox{MSE}(\cdot,\cdot)$ denotes the mean squared error, and $\mbox{MSA}(\cdot)$ denotes Multi-head Self Attention \cite{msa}.
\subsection{Adaptable Prompting Teacher}
Considering that the teacher is unaware of the student's capacity during training, an intuitive solution is to allow the teacher to optimize for distillation. However, the immense modelling capacity gap between the teacher and student degenerates the distillation into inferior self-distillation \cite{switchable}. Therefore, we propose inserting prompts into teacher, enabling it to manage distillation-specific knowledge while preserving the model's discriminative capacity. In Sec. \ref{section:numofprompt}, we find that the number of learnable prompts determine the discriminative capacities of teacher and student. For a plain Swin Transformer \cite{swin} with $K$ basic layers, an input facial image is resized and divided into $m$ patches $\mathbf{x}=\{x_j\in \mathbb{R}^{h\times w\times 3}\, |\,1\le j\le m\}$, $h$ and $w$ indicate the height and width of the image patches. Each patch is initially embedded into a patch feature:
\begin{equation}
\tilde{x}_j = Linear(x_j),\,\,\, \tilde{x}_j \in \mathbb{R}^d.
\label{eq1}
\end{equation}
Let $\tilde{\mathbf{x}}_0=\{\tilde{x}_j\, | 1\le j\le m\}$ denote the patch embedding set which refers to the input of $0$-th Transformer basic layer. We supplement a collection of learnable embeddings $\mathbf{p}$, initialized normally as prompts, into the embedding set $\tilde{\mathbf{x}}$. Let $t$ indicate the number of introduced prompts, which controls the adaptable capacity of the teacher, as detailed in Sec. \ref{section:numofprompt}. The Transformer backbone is initialized with a pre-trained model and remains frozen. During distillation, only the prompts specific to the distillation are optimized. Prompts are inserted exclusively into each Transformer basic layer. The prompts supplemented in the $i$-th Transformer basic layer is a set of d-dimensional vectors, denoted as $\mathbf{p}_i = \{p_j^i \in \mathbb{R}^{d} \,|1\le j\le t\}$. The feed-forwarding process of APT is formulated as follows:
\begin{equation}
\begin{aligned}
&[\hat{\mathbf{x}}_{i+1}, \hat{\mathbf{p}}_{i+1}] = B_i([\tilde{\mathbf{x}}_i, \mathbf{p}_i]),\\
&[\tilde{\mathbf{x}}_{i+1}] = PM([\hat{\mathbf{x}}_{i+1}]),\\
&[\hat{\mathbf{x}}_{i+2}, \hat{\mathbf{p}}_{i+2}] = B_{i+1}([\tilde{\mathbf{x}}_{i+1}, \mathbf{p}_{i+1}]),\\
&[\bm{f}] = [\hat{\mathbf{x}}_{K}, \hat{\mathbf{p}}_{K}].
\end{aligned}
\label{eq2}
\end{equation}
Here, $[\cdot\,,\cdot]$ denotes stacking and concatenation on the token dimension, and $B_i$ indicates the $i$-th Transformer basic layer. $\hat{\mathbf{x}}_{i+1}$ and $\hat{\mathbf{p}}_{i+1}$ refer to the output of patch tokens and prompt tokens from the $i$-th basic layer, respectively. $PM(\cdot)$ indicates the Patch Merging manipulation. We incorporate prompts as basic components in calculating shifted window attention \cite{swin} while ignoring them in the patch merging. In the subsequent layer, fresh prompts $\mathbf{p}_{i+1}$ are initialized and inserted in to the $\tilde{\mathbf{x}}_{i+1}$, as the input of $(i+1)$-th basic layer. Finally, we stack and concatenate the output of $K$-th layer as the pixel features $\bm{f}$ for both teacher and student.
\subsection{Unified Receptive Fields Mapping}
The purpose of URFM module is to map the pixel features extracted by the backbone into local features with unified receptive fields. Self-Attention (SA) can be considered an alternative solution due to its sequence-to-sequence functional form. However, it has two problems in our context. 1) SA maintains an equal number of input and output tokens, but the teacher and student networks generally have different numbers of pixel features (tokens), hindering the alignment of the feature space due to the inconsistent feature dimension between teacher and student. 2) The vanilla positional encoding method in vision \cite{ape} merely considers the spatial distance between tokens while disregarding the variation of face structure between tokens. The proposed URFM solves these problems by modifying the SA with 1) learnable query embeddings and 2) facial positional encoding.
First, we review the generic attention formulation containing query $\bm{Q}$, key $\bm{K}$, and value $\bm{V}$ embeddings.
\begin{equation}
\mbox{SA}(\bm{Q}, \mathbf{K}, \bm{V}) = Softmax\,(\frac{(\bm{Q}\bm{W}_{q}) (\bm{K}\bm{W}_{k})^\top}{\sqrt{d}}) \cdot \bm{W}_{v}\bm{V},
\label{eq3}
\end{equation}
where $\bm{W}_{q}$, $\bm{W}_{k}$, $\bm{W}_{v}$ are learnable weights and $d$ indicates the channel dimension. Then, we modify it with learnable local centers $\bm{C}$:
\begin{equation}
\mbox{SA}^{\prime}(\bm{Q}, \mathbf{K}, \bm{V}) = Softmax\,(\frac{(\bm{C}\bm{W}_{q}) (\bm{K}\bm{W}_{k})^\top}{\sqrt{d}}) \cdot\bm{W}_{v}\bm{V}.
\label{eq4}
\end{equation}
The local centers $\bm{C} \in \mathbb{R}^{L\times d}$ ensure consistent numbers of the output pixel features of both teacher and student networks.

Transformers inherently fails to capture the ordering of input tokens, which necessitates incorporating explicit position information through positional encoding (PE). The original visual transformer proposes inserting fixed encodings generated by sine and cosine functions of varying frequencies and learnable PE into the input \cite{vit}. Swin \cite{swin} supplements PE in the attention map as a bias, resulting in significant performance improvements, as formulated below:
\begin{equation}
e_{ij} = \frac{(\bm{f}_i\bm{W}_{q})(\bm{f}_j\bm{W}_{k})^\top+b_{ij}}{\sqrt{d}},
\label{eq5}
\end{equation}
where $e_{ij}$ represents the inner product between patch $i$ and $j$. $\bm{f}_i$ and $\bm{f}_j$ indicate the input elements of the patch embeddings. $b_{ij}$ refers to the learnable position weights indexed by the spatial distance between patches $i$ and $j$ \cite{swin}. However, the number of $Q$ and $K$ may not be identical in our setting, leading to inconsistent dimensions between the attention map and the patch distance matrix. To address this, we propose incorporating absolute PE for the query:
\begin{equation}
e_{ij} = \frac{(\bm{f}_i\bm{W}_{q}+b_{i})(\bm{f}_j\bm{W}_{k})^\top}{\sqrt{d}},
\label{eq6}
\end{equation}
where $b_i$ indicates the parameters indexed by the absolute position of patch $i$, which is formulated as follows:
\begin{equation}
b_{i} = P[I(\mbox{D}(i,anchor))].
\label{eq7}
\end{equation}
The index function $I(\cdot)$ maps a relative distance to an integer in a finite set, and we employ PIF \cite{ape} as the index function. $P$ is random initialized parameter buckets. $D(i,anchor)$ denotes the distance between patch $i$ and $anchor$. The conventional method for determining patch position involves measuring the Euclidean distance of coordinates from the anchor point \cite{ape}:
\begin{equation}
\tilde{\mbox{D}}(i,anchor) = \sqrt{(\hat{x}_i-\hat{x}_{anchor})^2+(\hat{y}_i-\hat{y}_{anchor})^2}.
\label{eq8}
\end{equation}
Let $(\hat{x}_i,\hat{y}_i)$ denote the coordinates of patch $i$. We choose $(\lfloor\!\!\!\;\frac{\sqrt{L}}{2}\!\rfloor,\lfloor\!\!\!\;\frac{\sqrt{L}}{2}\!\rfloor)$ patch as the anchor. $L$ indicates the number of local centers, as shown in Fig. \ref{fig3}. However, general visual PE methods primarily focus on the spatial distance of patch embeddings and overlook the differences in facial structure, which are pivotal in face domains. To address this, we propose incorporating facial structure distance into positional encoding, as formulated below:
\begin{equation}
\begin{aligned}
\mbox{D}(i,anchor) = \tilde{\mbox{D}}(i,anchor)+\gamma \cdot \mbox{D}_{face}(i,anchor).
\label{eq9}
\end{aligned}
\end{equation}
We define $\mbox{D}_{face}(i,j)$ as the facial structure distance between patch $i$ and $j$, and candidates are detailed in Sec. \ref{section:candidates}. $\gamma$ is a constant that unifies the range of spatial distance and facial structure distance.
\subsection{Facial Structure Distance}
\label{section:candidates}
In this section, we introduce two candidate methods for measuring the facial structure distance between patches, while preserving the Euclidean distance of coordinates for basic positional information.
\paragraph{Saliency Distance}
We utilize FaceX-Zoo \cite{facex} to predict 106 keypoints for each face image and quantify the saliency of each patch embedding by the amount of the keypoints inside the patch. The saliency distance between patches is computed as follows:
\begin{equation}
\mbox{D}_{face}(i,anchor)= \frac{|l_i-l_{anchor}|}{l_{max}},
\label{eq10}
\end{equation}
where $l_i$ and $l_{anchor}$ indicate the number of landmarks in the corresponding patch. $l_{max}$ denotes the maximum landmarks in a patch.
\paragraph{Relative Distance}
We use dlib \cite{dlib} to predict 5 keypoints for each face image and calculate the distance between the coordinates of the centroid of the patch $i$ and that of each keypoint to comprise a relative distance vector $\bm{d_i}=\{d_i^j, 1\le j\le5\}$, which is employed to determine the distance between patch $i$ and $anchor$:
\begin{equation}
\begin{aligned}
d_i^j = &\sqrt{|\bar{x}_i- x^l_j|^2+|\bar{y}_i- y^l_j|^2},\\ 
\mbox{D}_{face}(i,&anchor)= \mbox{Cos}\,(\bm{d}_i,\bm{d}_{anchor}),
\end{aligned}
\end{equation}
where $(\bar{x}_i,\bar{y}_i)$ denotes the coordinates of the centroid of patch $i$, and $(x^l_j,y^l_j)$ indicates the coordinates of $j$-th keypoint. $\mbox{Cos}(\cdot,\cdot)$ represents the cosine distance computation.
\section{Experiments}
\subsection{Dataset}
\paragraph{Training set}
For fair comparisons with other SOTA approaches, we employ the refined MS1MV2 \cite{arcface} as our training set. MS1MV2 consists of 5.8M facial images of 85K individuals.
\paragraph{Testing set}
We evaluate our method on several popular face benchmarks, including LFW \cite{lfw}, CFP-FP \cite{cfp-fp}, CPLFW \cite{cplfw}, AgeDB \cite{agedb}, CALFW \cite{calfw}, IJB-B \cite{ijb-b}, and IJB-C \cite{ijb-c}. LFW is a popular face verification dataset containing 13,233 images of 5,749 individuals.  Cross-Age LFW (CALFW) and Cross-Pose LFW (CPLFW) databases are constructed based on the LFW database to emphasize similar-looking, cross-age, and cross-pose challenges. CFP-FP database is built to facilitate significant pose variation, and the AgeDB-30 database is a manually collected cross-age database. The IJB-B and IJB-C are two challenging public template-based benchmarks for face recognition. The IJB-B dataset contains 1,845 subjects with 21.8K still images and 55K frames from 7,011 videos. The IJB-C dataset is a further extension of IJB-B, which contains about 3,500 identities with 31,334 images and 11,7542 unconstrained video frames. MegaFace Challenge \cite{megaface} consists of the gallery set, including 1M images of 690K subjects, and the probe set, including 100K photos of 530 persons from FaceScrub. We adopt the protocol in \cite{arcface}.
\begin{table*}[t]
    \centering
\resizebox{\textwidth}{!}{%
    \begin{tabular}{@{}ccccccccccc@{}}
    \toprule
        \multirow{2}{*}{\textbf{\makecell{Method\\(Transformer:CNN)}}} & \textbf{IJB-C} & \textbf{IJB-B} & \multicolumn{2}{c}{\textbf{MegaFace}} & \textbf{LFW} & \textbf{CFP-FP} & \textbf{CPLFW} & \textbf{AgeDB-30} & \textbf{CALFW} \\ \cmidrule(r){2-10}
&1e-4 & 1e-4 &Id & Ver & ACC & ACC & ACC & ACC & ACC \\ \midrule 
        Swin-S (Tea.)& 97.05 & 95.51 & 98.86 & 99.02 & 99.81 & 97.90 & 93.33 & 98.01 & 96.03  \\ 
        MobileFaceNet (Stu.)& 89.13 & 87.07  &90.91&92.71& 99.52 & 91.66 & 87.93 & 95.82 & 95.12  \\ \midrule
        FitNet \cite{fitnet}& 90.50 & 87.83 & 90.67 & 91.75 & 99.42 & 91.30 & 87.81 & 94.46 & 93.55  \\ 
        KD \cite{kd} & 92.42  & 89.79  & 91.04 & 92.63 & 99.52 & 91.92 & 88.69 & 94.94 & 94.60  \\ 
        DarkRank \cite{darkrank} & 93.06 & 90.78  & 91.42 & 93.21 & 99.40 & 91.81 & 87.35 & 94.98 & 94.71  \\ 
        SP \cite{sp} & 93.01 & 90.85  & 91.65 & 93.64 & 99.45 & 92.90 & 88.68 & 95.93 & 94.93  \\ 
        CCKD \cite{cckd} & 93.10  & 91.01  & 91.22 & 93.42 & 99.50 & 92.50 & 88.13 & 94.90 & 95.11  \\
        RKD \cite{cckd} & 91.92 & 89.48  & 89.95 & 91.23 & 99.26 & 91.87 & 88.71 & 94.83 & 94.71  \\ 
        EKD \cite{ekd} & 92.97  & 90.75  & 91.50 & 93.53 & 99.53 & 93.55 & 89.81 & 95.71 & 95.01  \\ 
        CKD \cite{accv} & 92.66 & 91.19  & 91.43 & 93.46 & 99.51 & 92.45 & 88.53 & 95.34 & 94.52  \\ 
        GKD \cite{gkd} & 94.33  & 92.13  & 94.98 & 95.23 & 99.52 & 92.81 & 89.96 & 95.90 & 95.11  \\ 
        \textbf{Ours} & \textbf{94.40} & \textbf{92.48} &\textbf{95.37}&\textbf{96.32}& \textbf{99.61} & \textbf{94.63} & \textbf{91.14} & \textbf{97.20} & \textbf{95.83}  \\ \bottomrule
    \end{tabular}%
}
\caption{Comparison on benchmark datasets of state-of-the-art knowledge distillation methods with our method. For large-scale face benchmarks \cite{ijb-b, ijb-c}, TPR@FPR=1e-4 is reported. For MegaFace Challenge \cite{megaface} using FaceScrub as the probe set, “Id” refers to Rank-1, and "Ver" refers to TPR@FPR=1e-6. For five small datasets \cite{lfw,cfp-fp,cplfw,agedb,calfw}, 1:1 verification accuracy is reported.}
\vspace{-0.4cm}
\label{tab:sota}
\end{table*}
\subsection{Experimental Settings}
\paragraph{Data Processing.} The input facial images are cropped and resized to 112$\times$112 for CNNs and ViT. For Swin, we utilize bilinear interpolation to resize the image from $112\times$112 to 224$\times$224. Then images are normalized by subtracting 127.5 and dividing by 128. For the data augmentation, we adopt the random horizontal flip.
\paragraph{Training.} We utilize Swin-S and ViT-S as the teacher models that are trained by ArcFace \cite{arcface}. For the student, there are two groups of student backbone networks widely used in face recognition, one is the MobileFaceNet \cite{mobilefacenet} that is modified based on MobileNet \cite{mobilenetv3}, and the other is the IResNet \cite{arcface} which is adapted from ResNet \cite{resnet}. To show the generality of our method, we utilize different teacher-student configurations. We set the batch size to 128 for each GPU in all experiments, and train models on 8 NVIDIA Tesla V100 (32GB) GPUs. We apply the SGD optimization method and cosine learning rate decay \cite{deit} with 4 warmup epochs and 16 normal epochs. The momentum is set to 0.9, and the weight decay is 5e-4. For ArcFace loss, we follow the common setting with s = 64 and margin m = 0.5. 
\subsection{Comparison with SOTA Methods}
In this section, we compare our method with state-of-the-art knowledge distillation methods, e.g., KD \cite{kd}, FitNet \cite{fitnet}, DarkRank \cite{darkrank}, RKD \cite{rkd}, SP \cite{ekd}, CCKD \cite{cckd}, EKD \cite{ekd} and GKD \cite{gkd}. We also compare our method with specifically designed method CKD \cite{accv} for cross-architecture knowledge distillation. Since the existing KD methods do not conduct experiments under cross-architecture knowledge distillation scenario for face recognition, we reproduce them according to the settings the the original manuscripts. 
\subsubsection{Results on LFW, CFP-FP, CPLFW, AgeDB and CALFW}
Tab. \ref{tab:sota} compares the results of the proposed methods with those of SOTA competitors on five face benchmarks. The results indicate that the majority of knowledge distillation methods surpass training the student network from scratch. Relation-based methods excel in comparison to feature-based methods but fall short of the performance achieved by methods specifically designed for cross-architecture scenarios. In contrast, our method synchronizes the receptive fields of local features in teacher and student networks, ultimately outperforming all competitors on small facial testing sets.
\begin{figure}[t]
\vspace{-0.25cm}
\subfloat[ROC on IJB-B]{
\centering
\label{fig4:a}
\includegraphics[width=0.5\columnwidth]{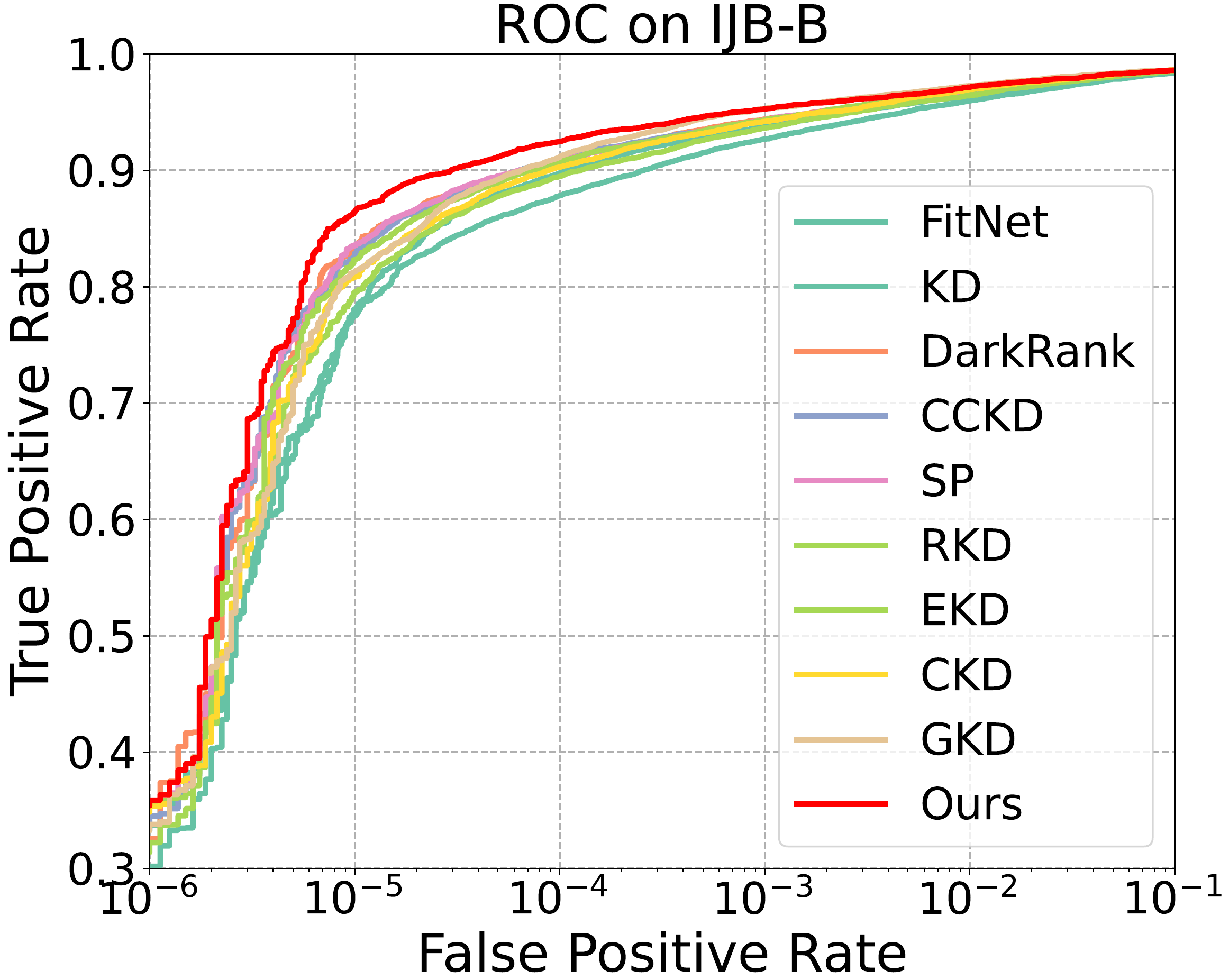}
}
\centering
\subfloat[ROC on IJB-C]{
\centering
\label{fig4:b}
\includegraphics[width=0.5\columnwidth]{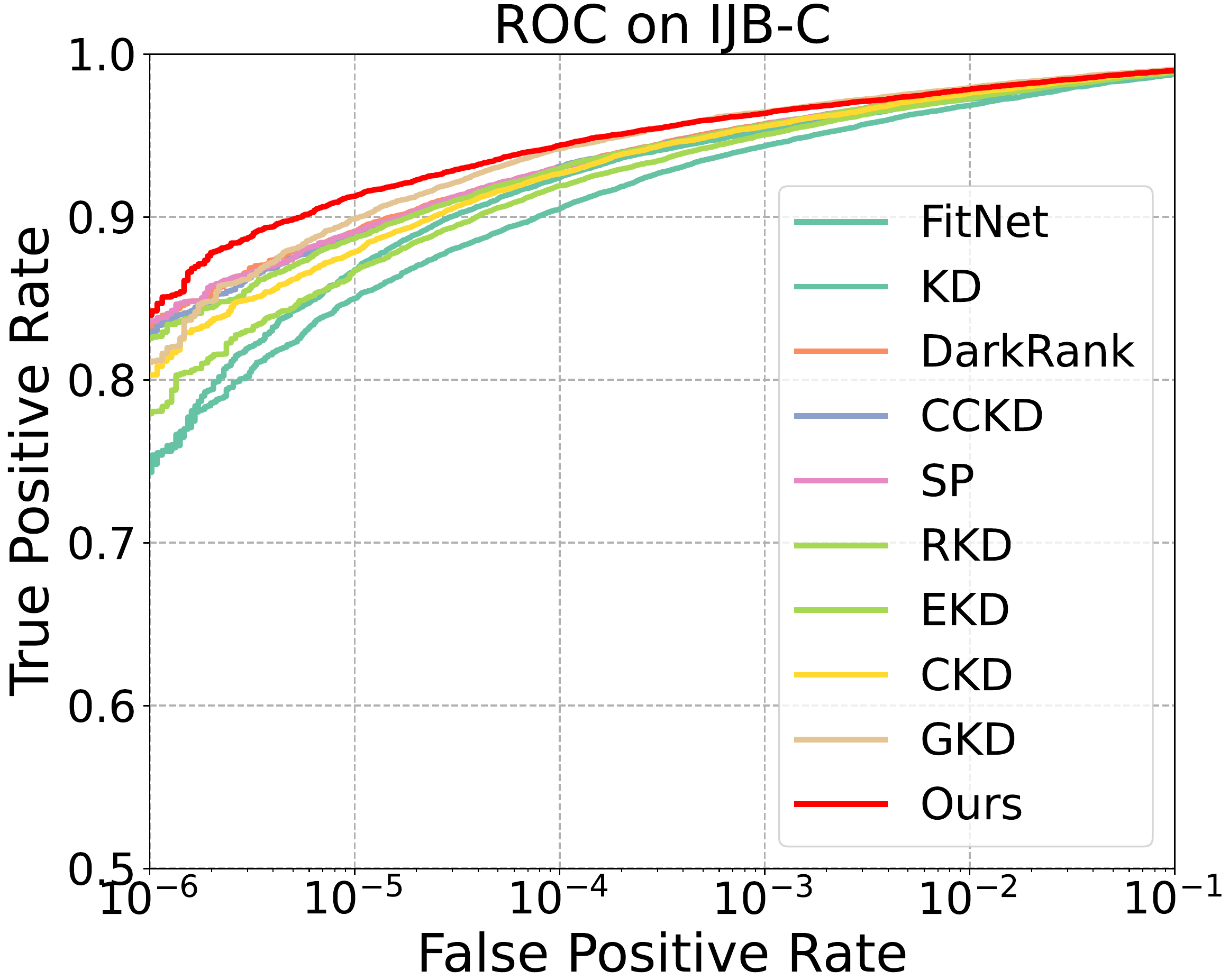}
}
\vspace{-0.1cm}
\caption{ROC curves of 1:1 verification protocol of different KD methods on the IJB-B and IJB-C dataset.}
\label{fig4}
\vspace{-0.5cm}
\end{figure}

\subsubsection{Results on IJB-B, IJB-C and MegaFace Challenge}
Tab. \ref{tab:sota} offers a comparison of the 1:1 verification TPR@FPR=1e-4 and TPR@FPR=1e-5 between existing state-of-the-art KD methods and the proposed method on IJB-B and IJB-C datasets. The majority of knowledge distillation methods exhibit substantial performance enhancements on these two large-scale datasets. Fig. \ref{fig4} presents the comprehensive ROC curves of current state-of-the-art competitors and our method, illustrating that our approach surpasses the other KD methods. For MegaFace challenge \cite{megaface}, we follow the testing protocol provided by ArcFace \cite{arcface}. As Tab. \ref{tab:sota} indicates, most competitors obtain superior performance than the baseline, whereas our method achieves the highest verification performance. For the rank-1 metric, our method performs marginally better than GKD \cite{gkd}.
\begin{table}[t]
    \centering
\resizebox{\columnwidth}{!}{%

    \begin{tabular}{@{}cccc|ccc@{}}

    \toprule
       \textbf{ASA} & \textbf{MA} & \textbf{APT} & \textbf{URFM} & \textbf{CFP-FP} & \textbf{CPLFW} & \textbf{AgeDB}   \\ \midrule
     \multicolumn{4}{c|}{Baseline} & 91.30 & 87.81 & 94.46 \\  \midrule
     $\bm{\checkmark}$ & & & & 92.45 & 88.53 & 95.34 \\  
     $\bm{\checkmark}$ & $\bm{\checkmark}$ & & & 92.72 & 89.05 & 95.85 \\  
     $\bm{\checkmark}$ & $\bm{\checkmark}$ & $\bm{\checkmark}$ & & 93.82 & 90.42 & 96.61 \\ 
     $\bm{\checkmark}$ & $\bm{\checkmark}$ & $\bm{\checkmark}$ & $\bm{\checkmark}$ & \textbf{94.63} & \textbf{91.14} & \textbf{97.20}  \\ \bottomrule 
    \end{tabular}%
}
\caption{Ablation experiments of Attention Space Alignment (ASA), Mutual alignment (Mutual), Adaptable Prompting Teacher network (APT) and Unified Receptive Fields Mapping (URFM). The baseline model is trained with FitNet \cite{fitnet}.}
\vspace{-0.3cm}
\label{tab:1}
\end{table}
\begin{table}[t]
    \centering
\vspace{-0.1cm}
\resizebox{\columnwidth}{!}{%
    \begin{tabular}{@{}cccccc@{}}
    \toprule
        \multicolumn{1}{c}{\textbf{Method}} & \textbf{CFP-FP} & \textbf{CPLFW} & \textbf{AgeDB-30} & \textbf{CALFW} \\ \midrule
        Euc \cite{ape} & 94.15 & 90.62 & 96.66 & 95.45      \\  \midrule 
	   Euc + RD & \textbf{94.78} & 90.85 & 96.87 & 95.71  \\ 
        Euc +  SD  & 94.63 & \textbf{91.14} & \textbf{97.20} & \textbf{95.83}  \\  \bottomrule 
    \end{tabular}%
}
\caption{Ablation of facial structural distance for indexing positional encoding. Experiments are conducted on basis of APT and URFM, and evaluated on popular facial benchmarks.}
\vspace{-0.3cm}
\label{tab:fpe}
\end{table}
\begin{table}[t]
    \centering
    \resizebox{\columnwidth}{!}{%
        \begin{tabular}{@{}cccccc@{}}
\toprule
            \multicolumn{1}{c}{\textbf{Method}} & \multicolumn{1}{c}{\textbf{CFP-FP}} & \textbf{CPLFW} & \textbf{AgeDB-30} & \textbf{CALFW} \\ \midrule
            $L=3\times 3$ & 93.04 & 89.86 & 95.68 & 94.56      \\ 
            $L=5\times 5$  & 94.42 &\textbf{91.15} & 96.93 & 95.58  \\  
            $L=7\times 7$  & \textbf{94.63} & 91.14 & \textbf{97.20} & \textbf{95.83}  \\ 
            \bottomrule
        \end{tabular}%
    }
    \caption{Ablation experiments of number of local centers. All experiments are evaluated on popular facial benchmarks.}
\vspace{-0.6cm}
    \label{tab:lc}
\end{table}

\subsection{Ablation Study}
\subsubsection{Effects of APT and URFM}
\label{section:ablation}
We employ FitNet \cite{fitnet} as our baseline model and conduct ablation experiments for Attention Space Alignment (ASA), Mutual Alignment via releasing teacher network parameters (MA), as well as the proposed Architecutre-Prompting Teacher network (APT) and Unified Receptive Fields Mapping (URFM), as shown in Tab. \ref{tab:1}. All the experiments are evaluated on popular face benchmarks, i.e., CFP-FP, CPLFW and AgeDB. The first and second rows show that aligning the attention spaces of the student and teacher networks results in a noteworthy performance enhancement. Comparing the second and third rows, we observe that allowing the teacher network to optimize for distillation enhances the student network's recognition performance. We argue that the tremendous modelling capability gap between the teacher and student networks degenerates the distillation to self-distillation, resulting in limited improvement. By contrast, APT confines the adaptable capacity of the teacher by introducing prompts in the network, thereby preserving the model's adaptable capacity and symmetric distillation. Note that we discard the final prompt embeddings to maintain an equal number of pixel features for teacher and student networks, i.e., $N^T=N^S$. Furthermore, we unify the pixel-wise receptive fields of the teacher and student through the URFM module, which further enhances the student's performance.

\subsubsection{Effects of Facial Structural Distance} We investigate two candidate facial structural distances for indexing positional encoding, i.e., Saliency Distance (SD) and Relative Distance (RD). We conduct the experiments based on APT and URFM, and compare the vanilla Euclidean distance (Euc) with SD and RD. From Tab. \ref{tab:fpe}, all methods outperform the baseline model with Euclidean positional encoding, and SD outperforms RD. Therefore, we choose SD as the metric of facial structural distance in the following experiments.

\subsubsection{Effects of Number of Local Centers} We investigate the effects of the number $L$ on local centers. We conduct the experiments after introducing APT and facial positional encoding. From Tab. \ref{tab:lc}, we can find $L=3\times3$ is inferior in comparison to others since few local centers hinder structural and spatial information of faces. In contrast, $L=7\times7$ achieves the best performance.
\begin{table}[t]
    \centering
\resizebox{\columnwidth}{!}{%
    \begin{tabular}{@{}cccccc@{}}
    \toprule
        \textbf{Method}  & \textbf{CFP-FP} & \textbf{CPLFW} & \textbf{AgeDB-30} & \textbf{CALFW} \\ \midrule
        Swin-S (Tea.) & 97.90 & 93.33 & 98.01 & 96.03  \\ 
        IR-18 (Stu.) & 94.60 & 89.97 & 97.33 & 95.70  \\ \midrule
        FitNet \cite{fitnet} & 94.08 & 90.03 & 96.58 & 95.23  \\ 
        CKD \cite{accv} & 94.11 &90.60 & 96.98 & 95.44  \\ 
        GKD \cite{gkd} & 94.85 & 91.01 & 97.58 & 95.75  \\ 
        \textbf{Ours} & \textbf{95.58} & \textbf{91.78} & \textbf{97.67} & \textbf{95.98}  \\ \midrule \midrule
        ViT-S (Tea.)& 96.19 & 92.55 & 97.82 & 95.92  \\ 
        MobileFaceNet (Stu.) & 91.66 & 87.93 & 95.82 & 95.12  \\ \midrule
        FitNet \cite{fitnet}& 91.10 & 87.46 & 94.48 & 94.40  \\ 
        CKD \cite{accv} & 92.25 & 88.51 & 95.85 & 95.00  \\  
        GKD \cite{gkd} & 92.14 & 89.65 & 95.58 & 95.06  \\ 
        \textbf{Ours} & \textbf{94.60} & \textbf{91.05} & \textbf{97.28} & \textbf{95.85} \\
 \bottomrule
    \end{tabular}%
}
\caption{Generalization for different student and teacher networks, as well as the identification comparisons with other SOTA methods. Student (Stu.) and teacher (Tea.) networks are replaced by IResNet-18 \cite{arcface} and ViT \cite{vit}, respectively.}
\vspace{-0.8cm}
\label{tab:5}
\end{table}
\begin{figure*}[t]
\centering
\vspace{-0.35cm}
\subfloat[Before Receptive Fields Alignment]{
\centering
\label{fig5:a}
\includegraphics[width=0.49\linewidth]{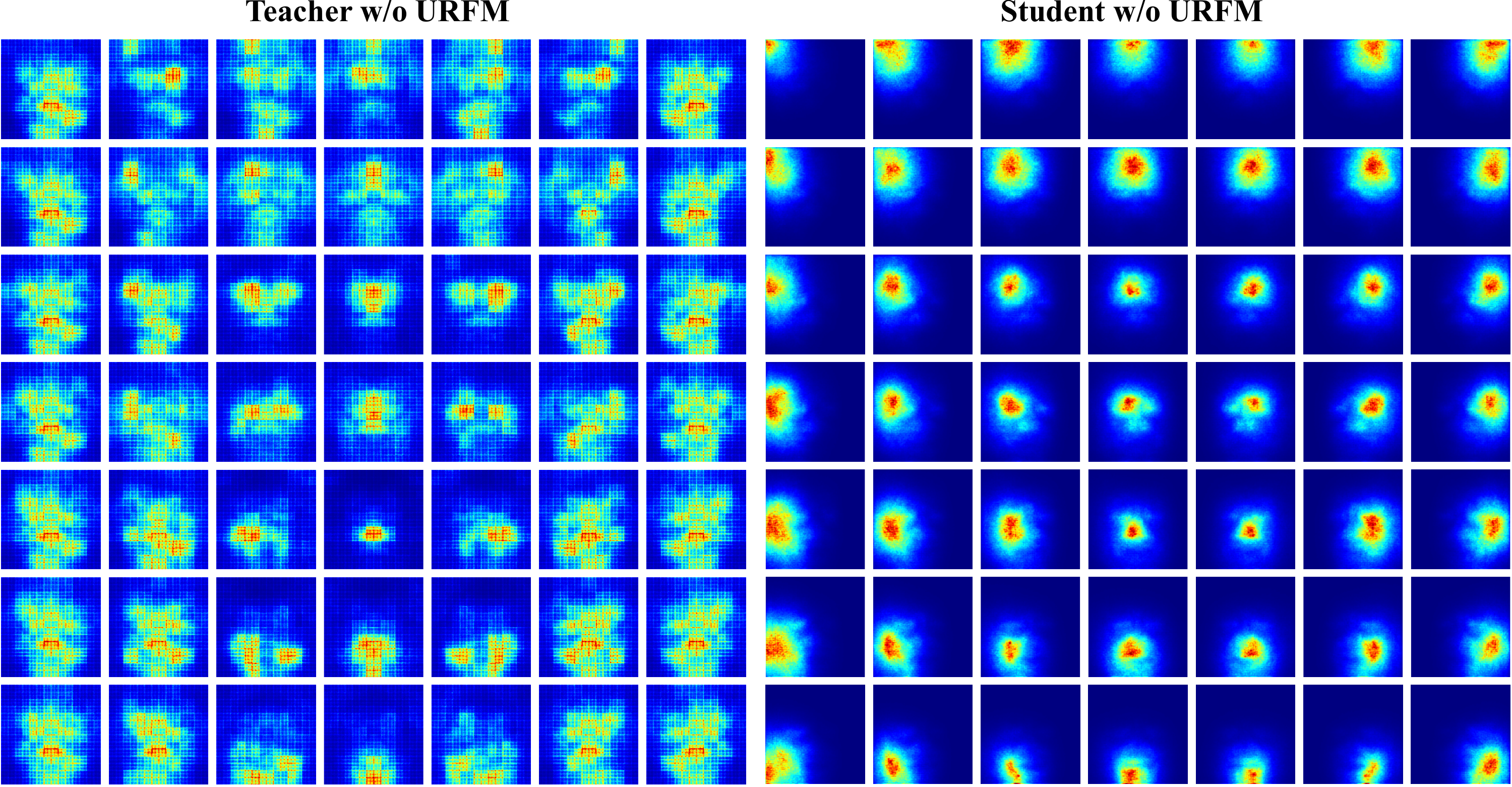}
}
\centering
\subfloat[After Receptive Fields Alignment]{
\centering
\label{fig5:b}
\includegraphics[width=0.49\linewidth]{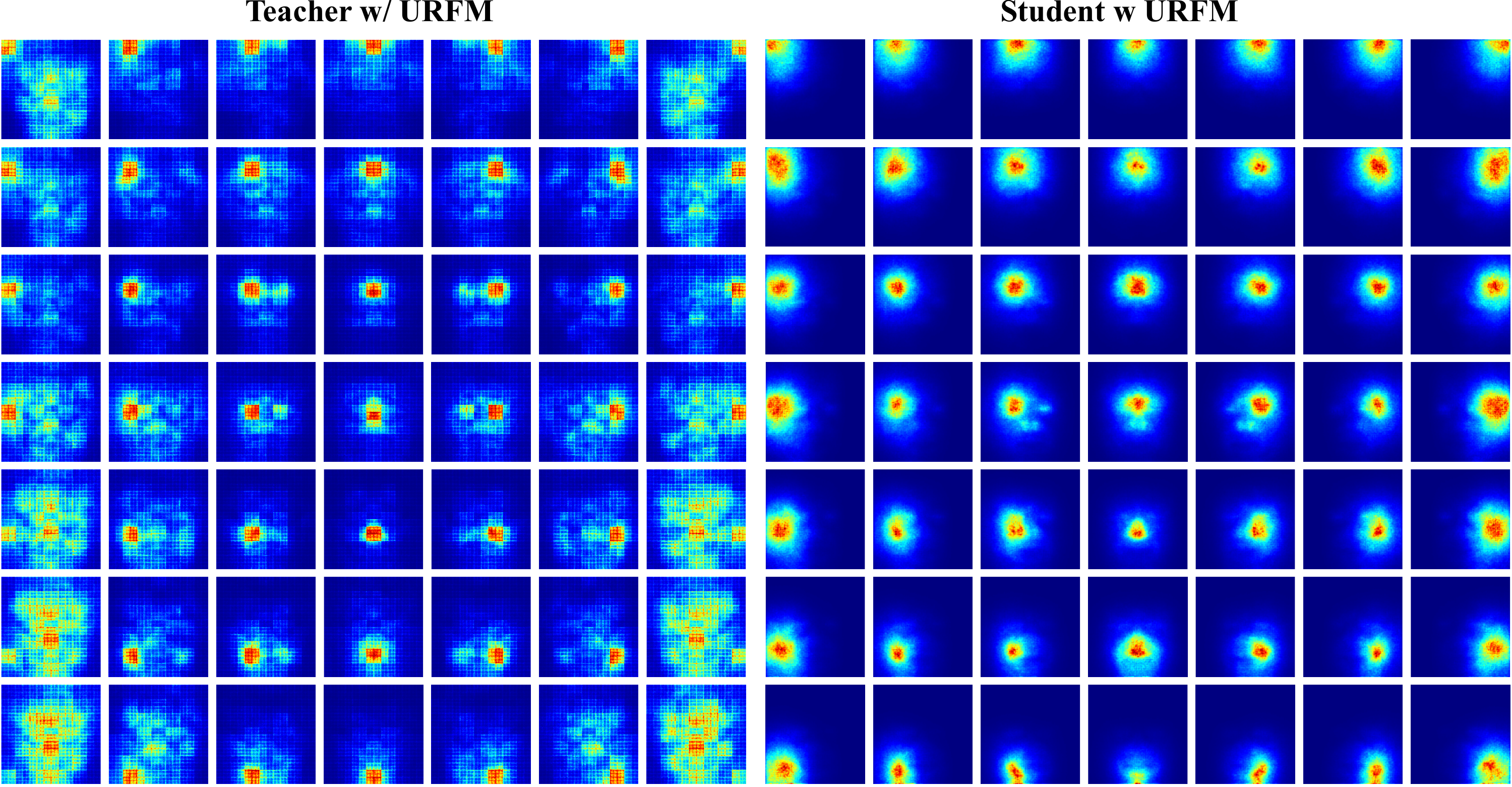}
}
\vspace{-0.3cm}
\caption{Pixel-wise Effective Receptive Fields (PERF) \cite{erf} before and after Unified Receptive Fields Alignment (URFM). We measure the PERF for teacher and student as the absolute value of the gradient of pixel features ($\bm{f}^T$ and $\bm{f}^S$) or the local features ($\bm{h}^T$ and $\bm{h}^S$). Results are averaged across all channels in the feature map for 500 randomly selected images.}
\label{fig5}
\end{figure*}

\subsubsection{Generalization for Student of IResNet}
We investigate the generalization of our method for the student of IResNet-18. As shown in Tab. \ref{tab:5}, the identification performance of different knowledge distillation methods is evaluated on CFP-FP, CPLFW, AgeDB and CALFW. Most methods outperform the baseline model on average performance, and our method improves the baseline and achieves the best performance on these four datasets.

\subsubsection{Generalization for Teacher of ViT}
To demonstrate the generalization of our method for different teacher networks, we select another Transformer branch, e.g., ViT \cite{vit}, as the teacher network. Following the settings in \cite{facetransformer}, we train a ViT-S and reproduce several SOTA KD methods. For simplicity, the URFM is fed with both patch embeddings and class tokens. As shown in Tab. \ref{tab:5}, most methods outperform the student trained from scratch with limited performance improvement. In contrast, our approach achieves superior performance over other methods.

\begin{figure}[t]
\captionsetup[subfigure]{labelformat=empty}
\vspace{-0.45cm}
\subfloat[All-learnable]{
\label{fig6:a}
\includegraphics[width=0.17\columnwidth]{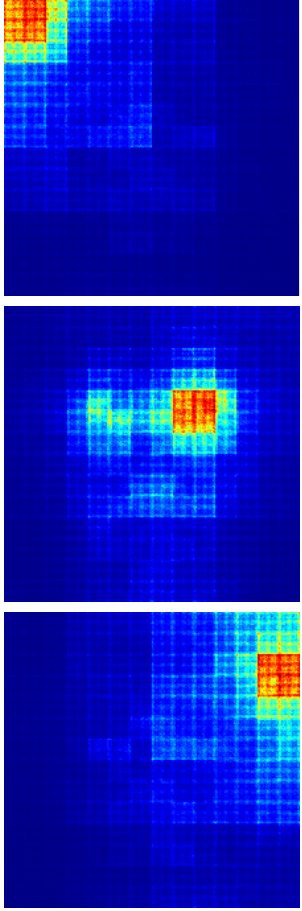}
}
\subfloat[50 Prompts]{
\centering
\label{fig6:b}
\includegraphics[width=0.17\columnwidth]{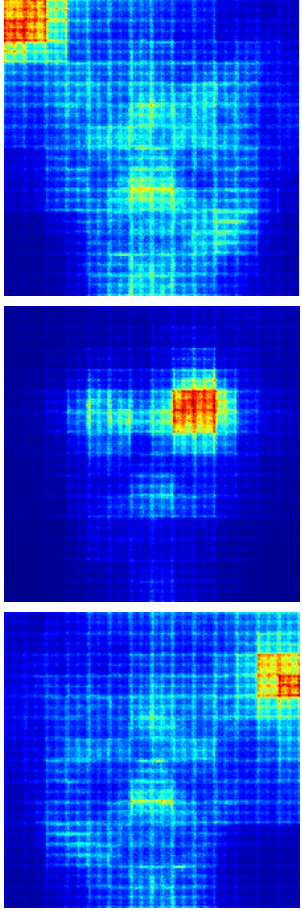}
}
\subfloat[25 Prompts]{
\centering
\label{fig6:c}
\includegraphics[width=0.17\columnwidth]{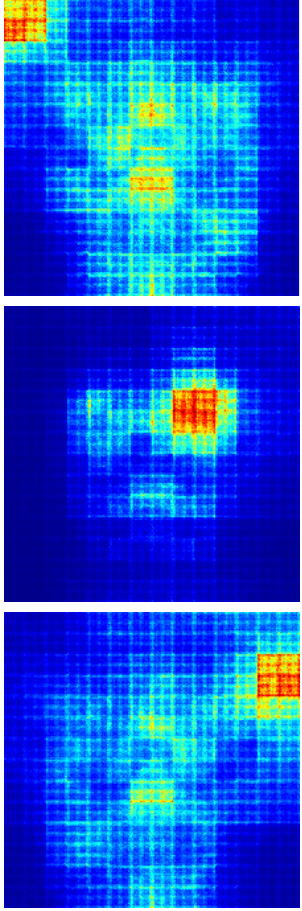}
}
\subfloat[5 Prompts]{
\centering
\label{fig6:d}
\includegraphics[width=0.17\columnwidth]{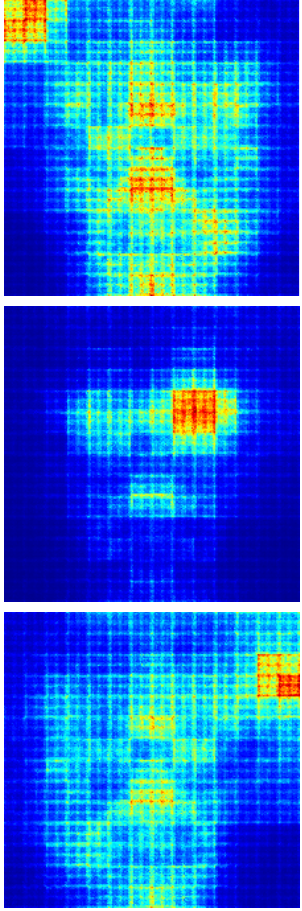}
}
\centering
\subfloat[Frozen]{
\centering
\label{fig6:e}
\includegraphics[width=0.17\columnwidth]{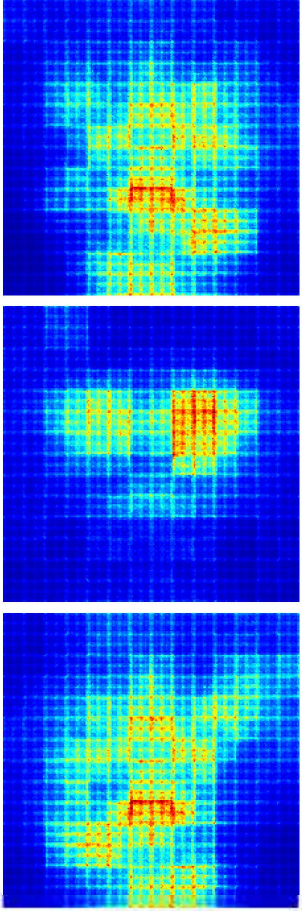}
}
\vspace{-0.3cm}
\caption{Pixel-wise effective receptive fields \cite{erf} of different teachers with various adaptable capacity degrees, which is adapted by altering learnable parameters in teacher.}
\label{fig6}
\vspace{-0.45cm}
\end{figure}
\subsection{Analysis}
\subsubsection{Pixel-wise Receptive Fields Alignment}
To showcase the alignment of pixel-wise receptive fields between the teacher and student, we compute and visualize the Effective Receptive Field (ERF) \cite{erf} of their local features, denoted as PERF. Fig. \ref{fig5} shows that URFM aligns the PERF of the teacher with that of the student, whereas the PERF of the student undergoes minimal change. Additionally, we find the ERF of the Transformer exhibits a grid-like pattern.
\subsubsection{Adapting Teacher's Adaptable Capacity}
The teacher's adaptable capacity is manifested in the degree of alignment in PERF of teacher and student. We analyze that the scale of learnable parameters can adapt the teacher's adaptable capacity. As depicted in Fig. \ref{fig6}, we alter the number of incorporated prompts in the teacher and visualize the resulting PERF. We find that the PERF of the teacher converges with that of the student as the number of prompts increases, thereby demonstrating that the number of prompts is able to reflect the teacher's adaptable capacity. Notably, we consider that the teacher without prompt and optimizing all parameters in distillation hold the highest adaptable capacity, whereas the teacher frozen in distillation exhibits the lowest adaptable capacity, denoted as "All-learnable" and "Frozen", respectively.
\subsubsection{Trade-off between Teacher's Adaptable Capacity and Discriminative Capacity}
We first explore the effects of the teacher's adaptable capacity on its performance in CPLFW and AgeDB. Tab. \ref{tab:numofprompt} reveals a decline in the teacher's discriminative capacity as the adaptable capacity increases, since higher adaptable capacity results in overfitting in the distillation. To identify the easy-to-learn teacher, we evaluate the performance of the students distilled by teachers with different degrees of adaptable and discriminative capacity. As shown in Tab. \ref{tab:numofprompt}, we find that the teacher with the highest discriminative capacity and lowest adaptable capacity ("\textcolor{darkgreen}{Frozen}") is hard-to-learn. In contrast, the teacher with the lowest discriminative capacity but the highest adaptable capacity ("\textcolor{dblue}{All-learnable}") exhibits a marginal improvement. Interestingly, we observe that the teacher with a trade-off between discriminative capacity and adaptable capacity yields optimal performance for the student.
\label{section:numofprompt}
\begin{table}[t]
\vspace{-0.1cm}
\renewcommand\arraystretch{1.4}
    \centering
    \resizebox{\columnwidth}{!}{%
        \begin{tabular}{@{}cccc|cc@{}}
\hline
            \multicolumn{4}{c|}{\textbf{Teacher}} &  \multicolumn{2}{c}{\textbf{Student}}\\ \hline
            \textbf{Adaptable Capacity} & \multicolumn{1}{|c}{\textbf{Learnable Params}} & \multicolumn{4}{|c}{\textbf{Performance}}   \\ \hline
           \textcolor{dblue}{Highest} & \textcolor{dblue}{All-learnable}  & \textcolor{darkgreen}{91.24} & \textcolor{darkgreen}{97.18} & 89.77 & 96.12 \\ \hline
           High & 50 Prompts  & 92.10 & 97.47 & 91.00 & 96.61 \\  
           Medium & 25 Prompts & 93.00 & 97.86 & \textcolor{navy}{\textbf{91.14}} & \textcolor{navy}{\textbf{97.20}} \\ 
 	Low & 5 Prompts & 93.18 & 97.95 & 90.86 & 96.68 \\ \hline
		\textcolor{darkgreen}{Lowest} & \textcolor{darkgreen}{Frozen} & \textcolor{dblue}{93.33} & \textcolor{dblue}{98.01} & 89.23 &  95.94 \\ 
            \hline
        \end{tabular}%
    }
    \caption{Effects of varying adaptable and discriminative capacity (performance) of the teacher on student's performance. \textcolor{dblue}{Blue} and \textcolor{darkgreen}{Green} denote the highest and lowest adaptable or discriminative capacity, respectively. \textcolor{navy}{Red} denotes the trade-off that results in the best performance for the student. }
    \label{tab:numofprompt}
\vspace{-0.9cm}
\end{table}

\section{Conclusion}
In this paper, we first demonstrate the implication of the architecture gap in cross-architecture knowledge distillation for face recognition. Subsequently, we find two challenges for CAKD in face recognition: 1) the teacher and student share disparate spatial information for each pixel, obstructing the alignment of feature space, and 2) the teacher network is not trained in the role of a teacher, lacking proficiency in handling distillation-specific knowledge. To tackle these problems, 1) we present a Unified Receptive Fields Mapping module (URFM), aiming at mapping pixel features of the teacher and student into local features with unified receptive fields. Additionally, 2) we propose an Adaptable Prompting Teacher network (APT) that supplements an adaptable number of prompts into the teacher network to instruct the network to manage distillation-specific knowledge. We experimentally find that the teacher with a trade-off between discriminative capacity and adaptable capacity is the most easy-to-learn for the student. Moreover, we construct different teacher-student pairs and demonstrate the generalization of the proposed method to different network settings. Finally, extensive experiments on popular face benchmarks and two large-scale verification datasets demonstrate the superiority of our method.

\bibliographystyle{ACM-Reference-Format}
\bibliography{sample-base}


\end{document}